\documentclass[10pt, conference,letterpaper]{IEEEtran}
\IEEEoverridecommandlockouts
\usepackage{cite}
\usepackage{amsmath,amssymb,amsfonts,amsthm}
\usepackage{graphicx}

\usepackage{textcomp}
\usepackage{xcolor}
\usepackage{bm}
\usepackage{comment}
\usepackage{booktabs}
\usepackage{multirow}
\usepackage{multicol}

\DeclareMathOperator*{\argmax}{\arg\!\max} 

\newcommand{\marco}[1]{{[\textcolor{red}{\bf Marco's Note: #1}]}}

\usepackage{algorithm}
\usepackage{algcompatible}

\usepackage[caption=false]{subfig}

\makeatletter
 \let\old@ps@headings\ps@headings
 \let\old@ps@IEEEtitlepagestyle\ps@IEEEtitlepagestyle
 \def\confheader#1{%
 \def\ps@headings{%
 \old@ps@headings%
 \def\@oddhead{\strut\hfill#1\hfill\strut}%
 \def\@evenhead{\strut\hfill#1\hfill\strut}%
 }%
 \def\ps@IEEEtitlepagestyle{%
 \old@ps@IEEEtitlepagestyle%
 \def\@oddhead{\strut\hfill#1\hfill\strut}%
 \def\@evenhead{\strut\hfill#1\hfill\strut}%
 }%
 \ps@headings%
 }
 \makeatother
\confheader{%
 Accepted for Presentation at the IEEE DySPAN 2021, ML Network
Automation and Control, ©2021 IEEE
 }

\begin{document}


\title{Federated Deep Reinforcement Learning for the Distributed Control of NextG Wireless Networks}

\author{\large Peyman Tehrani$^\dagger$, Francesco Restuccia$^*$  and Marco Levorato$^\dagger$\\
\normalsize $^\dagger$Donald Bren School of Information and Computer Sciences, University of California at Irvine, United States\\
  $^*$Department of Electrical and Computer Engineering, Northeastern University, United States\\
\normalsize e-mail: \{peymant,~levorato\}@uci.edu, frestuc@northeastern.edu
 \thanks{This work was partially supported by the NSF grants MLWiNS-2003237 and CNS-2134567.}
}

\maketitle

\begin{abstract}
Next Generation (NextG) networks are expected to support demanding tactile internet applications such as augmented reality and connected autonomous vehicles. Whereas recent innovations bring the promise of larger link capacity, their sensitivity to the environment and erratic performance defy traditional model-based control rationales. Zero-touch data-driven approaches can improve the ability of the network to adapt to the current operating conditions. Tools such as reinforcement learning (RL) algorithms can build optimal control policy solely based on a history of observations. Specifically, deep RL (DRL), which uses a deep neural network (DNN) as a predictor, has been shown to achieve good performance even in complex environments and with high dimensional inputs. However, the training of DRL models require a large amount of data, which may limit its adaptability to ever-evolving statistics of the underlying environment. Moreover, wireless networks are inherently distributed systems, where centralized DRL approaches would require excessive data exchange, while fully distributed approaches may result in slower convergence rates and performance degradation. In this paper, to address these challenges, we propose a federated learning (FL) approach to DRL, which we refer to federated DRL (F-DRL), where base stations (BS) collaboratively train the embedded DNN by only sharing models' weights rather than training data. We evaluate two distinct versions of F-DRL, value and policy based, and show the superior performance they achieve compared to distributed and centralized DRL.

\end{abstract}

\begin{IEEEkeywords}
Deep reinforcement learning, Federated Learning, Power control, Multi agent reinforcement learning, Wireless networks, Resource allocation.
\end{IEEEkeywords}

\section{Introduction}

On the one hand, next generation (NextG) networks are expected to support a wide range of essential and demanding real-time services such as augmented reality, connected autonomous vehicles and in-network computing that require coherent performance. On the other hand, recent advancements at the physical layers such as millimeter Wave (mmW) communications, while empowering the network with increased capacity, make its temporal behavior more erratic and convoluted. The NextG network environment, then, presents inherent control challenges that defy traditional model-based control approaches. To address these daunting challenges, the \emph{zero-touch} network paradigm utilizes machine learning to eliminate the need for human-based design, and enable fast data-driven adaptation to different operating conditions.
 
Power and bandwidth allocation is one of the fundamental problems in wireless networks. While many optimization frameworks have been proposed in this domain \cite{tehrani2016resource}, by directly utilizing the data originated by the system, data-driven methodologies have the potential to improve the adaptability of the network to environmental and traffic conditions and, ultimately, improve performance. Moreover, they are inherently more robust compared to model-based ones, as the latter class may suffer from model mismatch in real-world settings. 

Among data driven algorithms, deep reinforcement learning (DRL) has achieved state-of-the-art performance in high dimensional complex control problems \cite{mnih2015human}. In DRL frameworks, the agent iteratively interacts with
the environment to learn the optimal control policy. Additionally, DRL is often much faster in selecting the optimal action compared to  conventional optimization methods which usually requires iterative computation and matrix inversion. 

Most of the previous works in this domain proposed either fully centralized \cite{khan2020centralized,meng2020power} or  distributed \cite{sinan2020deep,zhang2019calibrated} DRL algorithms to solve the resource allocation problem. In the former case, training and decision making is centralized, where all BSs send all state information at every time step to the server. In the latter case, each BS independently trains and executes the model without sharing information with other BSs. 

In this paper, we propose a \emph{federated} implementation of DRL algorithms which takes the advantages of both distributed and centralized solutions. Federated Learning (FL) \cite{mcmahan2017communication} is a family of machine learning problems where many clients (\emph{e.g.}, mobile devices or whole organizations) collaboratively train a DNN model under the orchestration of a central server (\emph{e.g.}, a service provider), while keeping the training procedure decentralized. FL embodies the principles of focused collection and data minimization, and can mitigate many of the systemic privacy risks and costs resulting from traditional, centralized, training of machine learning models. This area has received considerable recent interest, both from academy and industry. However, so far FL have mainly been applied to supervised learning problems in fields such as NLP and machine vision, and there have been few contributions  that used federated learning to train distributed control and DRL models. In this domain, a federated control approach has been proposed in \cite{kumar2017federated} to solve coordination problems among a multitude of agents, edge caching problems \cite{wang2020attention,wang2020federated} and mobile edge computing strategies and offloading \cite{zhu2020federated,ren2019federated} and other internet of things (IoT) applications\cite{nguyen2021federated}.

Here we consider the maximization of the downlink sum rate  of a multi-cell network with mobile users. The base stations (BSs) have access to local information and collaborate with each other by sharing their model weights in a FL fashion to quickly train the DNN model at the core of the DRL controller. We will consider different value-based and policy-based DRL algorithms and compare the performance and efficiency of distributed and federated versions of these algorithm. We demonstrate that our F-DRL approach improves the performance in terms of overall network sum rate compared to fully distributed implementations by more than $40 \%$. Our F-DRL approach also reduces communication overhead between BSs and the central server compared to fully centralized solutions and also only need to share the model weights publicly so would protect the local users information privacy in each cells. 
\section{Literature Review}
Recent work proposed the use of DRL algorithms to solve a wide range of optimization problems in wireless communication networks. Power allocation is one the main areas, with several contributions using DRL to find the optimal power \cite{zhang2018power,meng2019power,meng2020power,nasir2019multi,sinan2020deep,ding2020deep}. Some of these contributions use deep Q-learning on discrete power levels \cite{meng2019power,meng2020power,saeidian2020downlink} and other states of the art DRL algorithms, such as deep deterministic policy gradient (DDPG) \cite{sinan2020deep,meng2020power} and trust region policy optimization (TRPO) \cite{khan2020centralized} on continuous power control in multi cell network scenarios.
In addition to power allocation, recent work explored the use of DRL for a wide range of resource management problems, including  spectrum allocation \cite{lei2020deep}, joint user association and resource allocation \cite{ding2020deep} and channel selection \cite{tan2020deep}.

In this paper, we expand on this exciting area by proposing a framework where multiple -- federated -- DRL agents collaboratively learn a shared predictive model, while all training data remain local to the corresponding device (that is, the BS). The key motivations to integrate FL with DRL are (\emph{i}) the boosting in training speed compared to fully distributed implementation, and (\emph{ii}) the reduced amount of data to be shared. Regarding the latter point, we note that centralized DRL approaches to this problem need to exchange data for real-time control, while in our F-DRL frameworks the agents exchange the model weights to update the individual models. Thus, the data exchange has a more relaxed delay constraint and the backbone network's load decreases.

\section{System Model and Problem Formulation}
We consider a cellular network, where $N$ different Base Stations (BS) serve $K$ mobile users. We assume that both the BSs and users are equipped with one
transmit antenna and each BS is deployed at the cell center. We index the BSs with $ n \in \mathcal{N}= \{1,2,...,N\}$ and the users with $ k \in \mathcal{K}= \{1,2,...,K\}$. We denote the channel gain between the $n$th BS and $k$th user in cell $j$ at time slot $t$ with:
\begin{equation}
 g^t_{n,j,k}=|h^t_{n,j,k}|^2\alpha_{n,j,k},
\end{equation}
where $h^t_{n,j,k}$ is the small scale fading factor with Rayleigh distributed envelope and $\alpha_{n,j,k}$ is the large-scale fading component, which includes path loss and log-normal shadowing. We model the small-scale Rayleigh fading component according to the Jakes fading model, that is, $h^t_{n,j,k}$  is assumed to be a first-order complex Gauss-Markov process:
\begin{equation}
 h^t_{n,j,k}=\rho h^{t-1}_{n,j,k}+\sqrt{1-\rho^2}e^{t}_{n,j,k}.
\end{equation}
Here, the innovation process variables $e^{t}_{n,j,k}$ are identically distributed circularly symmetric complex Gaussian random variables with unit variance, independent from $h^{t-1}_{n,j,k}$. The temporal correlation between two consecutive fading component is
\begin{equation}
\rho = J_0(2\pi f_d T_s ),
\end{equation}
 where $J_0(.)$ is the zeroth-order Bessel function of the first
kind, $f_d$ is the maximum Doppler frequency, and $T_s$ is the duration of one time slot. A higher mobility of users leads to a higher Doppler frequency and thus a lower temporal correlation of the channel.

Denoting the transmission power from BS $n$ to its user $k$ at slot $t$ with $p^{t}_{n,k}$, the downlink signal-to-interference-plus-noise ratio (SINR) of user $k$ in cell $n$ at time slot $t$ is:
\begin{equation}
 \gamma_{n,k}=\frac{p^{t}_{n,k} g^t_{n,j,k}}{I_i + I_o +N_{k}},
\end{equation}
where $N_k$ is the  noise power at user $k$ and $I_i$ and $I_o$ are the
intra-cell and inter-cell interference, respectively:
\begin{align}
    I_i & = \sum_{k' \neq k}  g^t_{n,n,k} p^{t}_{n,k'}, \\
    I_o & = \sum_{n' \neq n} g^t_{n',n,k} \sum_j p^{t}_{n',j}.
\end{align}
The data rate at user $k$ in cell $n$, then, is: 
\begin{equation}
 C_{n,k}=B \log(1+\gamma_{n,k}),
\end{equation}
where $B$ is the bandwidth available to the network.

Our goal is to find the set of downlink transmission powers $p^{t}_{n,k}$ that maximize the sum rate of the whole network under a constraint on the maximum power. Formally, the optimization problem is:
\begin{align}
&\max_{p^{t}_{n,k}} \sum_n\sum_k C_{n,k} \label{eq:optprob}\\
&\textrm{s.t.} \quad  0 \leqslant	 p^{t}_{n,k} \leqslant	 P_{max} \quad    \forall k,n, \nonumber
\end{align}
where $P_{max}$ is the the maximum transmission power for $k$th AP  and minimum data rate requirement at $k$th user, respectively.

Clearly, due to the interference terms in the denominator of SINRs, the optimization problem is non-convex and its solution non trivial. Importantly, non-convexity is not the only challenge to overcome to solve the problem. In fact, while iterative algorithms can be developed that achieve good performance, these algorithms require compute-intense operations such as matrix
inversion and bisection or singular value decomposition in
each iteration, which makes their real-time implementation
difficult. Additionally, these algorithms need full access to channel state information  (CSI) of all the users to derive the optimal solution. 

\begin{figure}
\centering
        \includegraphics[width=.99\columnwidth]{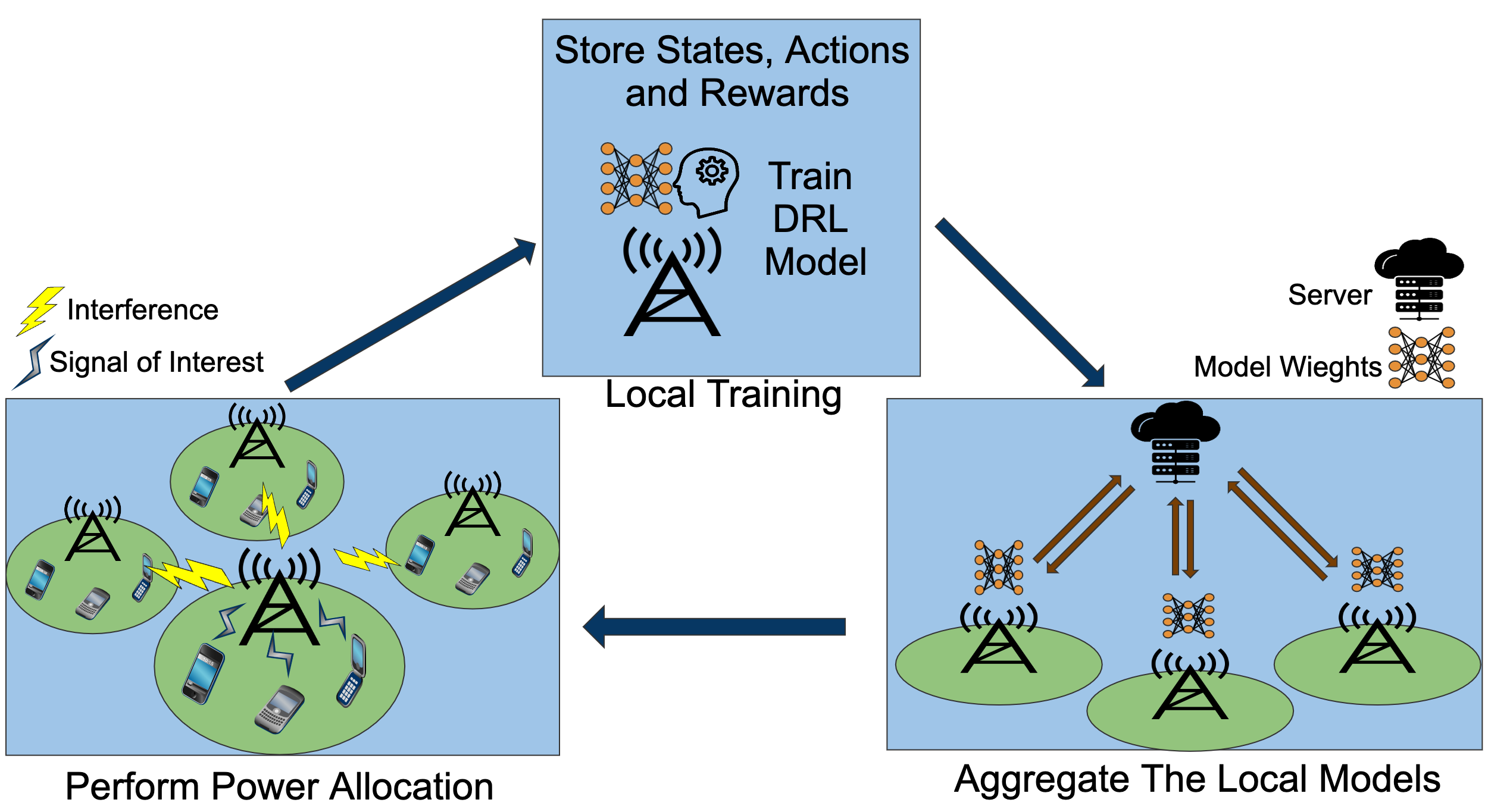}
        \caption{Overall view of the network model and  F-DRL procedure.}
        \label{fig:sys}
\end{figure}

Therefore, the resolution of the optimization problem requires a self-adaptive solution feasible for execution at run-time while achieving good performance having access only to partial observations of the environment. We, then, reformulate the problem as a multi agent RL (MARL) problem , where each access point is an agent which in each time slot determines the transmit power allocated to its associated users. Then, based on the feedback that it receives from the network  (which could be a function of rate and powers of other users and neighbor BSs), the BSs adapt their transmit power. Fig.~\ref{fig:sys} illustrates the framework we propose.

\section{Federated Deep Reinforcement Learning}

First, we redefine the problem (\ref{eq:optprob}) in a RL setting, where each BS is an agent whose objective is to maximize the sum rate of its own users, while mitigating interference to neighboring cells. Thus, each BS has separate control policies that output the optimal power levels given the current observed state. In the context of DRL, the base stations need to train DNN models whose output is either the Q-values or the control action (defined later). One of the critical issues, then, is training the DNN models as fast as possible to adapt to the current network conditions. In order to speed up training, we propose the federated deep reinforcement learning (FDRL) framework, where the federated agents -- the BSs -- collaboratively learn a predictive model by sharing their DRL model weights while keeping their users data private.

In the following sections, first we define the RL problem in terms of state space, action space and the reward function corresponding to problem (\ref{eq:optprob}). Then, we propose two versions of FDRL: federated deep Q network (FDQN) and federated deep policy gradient algorithm (FDPG) to solve the distributed power control problem.

\subsection{RL Formulation}

RL algorithms are usually set in the context of
Markov decision processes (MDP), defined by the 5-tuple $\langle \mathcal{S}, \mathcal{A},\mathcal{P}, \mathcal{R}, \gamma \rangle$, where 
$\mathcal{S}$ is the state space, $\mathcal{A}$ is the action space, $Pr(s_{t+1}
|s_t, a_t) \in \mathcal{P}$ are the transition probabilities, $r_t(s_t,a_t) \in \mathcal{R}$ is the reward function, and $\gamma \in \big[0, 1\big)$ is the discount factor.
At each time $t$, based on the current state $s_t$, the agent takes an
action $a_t \in \mathcal{A}$ and transitions from state $s_t$ to a new state $s_{t+1}$ with probability $Pr \big(s_{t+1}|s_t, a_t \big)$, receiving a reward $r_{t}$. We define the
policy $\pi\big(s, a\big)$ as the probability of taking action $a_t = a$ in
state $s_t = s$, that is, $\pi\big(s, a\big) = Pr\big(a_t {=} a|s_t {=} s\big)$. 

The goal of the RL agent is to learn a policy that maximizes the expected
sum of discounted rewards that it receives over the long run, also called return:
$R_t = \sum_{ i=0 }^{\infty} \gamma^{i+t} r_{t+i}$. We define the  optimal policy $\pi^*$ as the policy that maximizes the expected return from a state $s$:
\begin{equation} 
\pi^* = \argmax_{\pi} {E}_{\pi}\{ R_0|s_0=s \}.
\end{equation}
Here we define the states, actions and reward function for our power control problem.

\subsubsection{States}

The main features that describe the network system are the channel gains between the BSs and users, and the previous transmission power and rate. We assume BSs have only access to information on neighboring cells in the set $U_n$. The feature set we include in the state of BS $n$, then, is:
\begin{equation}
\mathcal{S}=\big\{g^t_{n,j,k},p^{t-1}_{n,k},C^{t-1}_{n,k}\big\}  \quad    \forall j,k \in U_n \\
\end{equation}

\subsubsection{Actions}
We use discrete power levels which take values between $0$ and $P_{max}$ defined as
\begin{equation}
\mathcal{A}=\left\{0,\frac{P_{max}}{M-1},\frac{2P_{max}}{M-1}, ...., P_{max}\right\},
\end{equation}
where $M$ is the number of power levels. All agents have the same action space.

\subsubsection{Reward Function}
In order to design the appropriate reward function, we need
to estimate the progress of the $n$th BS toward the goals of
the optimization problem (\ref{eq:optprob}). To this aim, the reward function considers both the sum-rate of the agent's cell and the interference generated to the neighbor cells $U_n$. We, then, define the reward function as:
\begin{equation}
r_t=\sum_k C^t_{n,k}+\beta\sum_{n' \in U_n}\sum_k C^t_{n',k},
\end{equation}
where here $\beta$ is a parameter controlling the tradeoff between sum-rate and interference. If $\beta$ is set to $0$, then the agent ignores the performance degradation caused to other cells (which in turn do the same). As $\beta$ increases, the BSs take a conservative stance, that is, they privilege interference reduction with respect to their own achieved sum-rate.

\subsection{Federated Learning Formulation}

As described later in this section, DRL algorithms use DNNs to produce either Q-values or probability of taking an actions to optimize the return.
The canonical federated learning problem involves learning a single, global statistical model (in our case a DNN) from data produced by a multitude of devices. In our framework, we learn this model under the constraint that in each of the BSs state information is stored and processed locally, with only intermediate model updates being communicated periodically to a central server. The goal of the training is to minimize the following objective function:
\begin{equation}\label{eq:fed}
 \min_{\theta} F(\theta)=\sum_{i=1}^N w_i F_i(\theta_i), 
\end{equation}
where $F(\theta)$ and $\theta$ are the global function and global model weights to be optimized and $F_i$ and $\theta_i$ are the local loss function and local model weights at BS $i$, and $w_i$ is the contribution of cell $i$ to whole network in terms of data size which is equal to $w_i=\frac{k_i}{\sum_{i=1}^N k_i}$ where $k_i$ is the number of users in cell $i$ .

\subsection{Federated Deep Q Network}

Formally, value-based RL algorithms use an action-value function $Q(s, a)$ to estimate an expected return starting from state $s$ when take action $a$:
\begin{align}
 Q_{\pi}(s_t,a) & =  E_{\pi}\{ \sum_{ k=1 }^{\infty} \gamma^{k-1} r_{t+k-1}|s_t,a\} \\
                & = E_{s_{t+1},a}\{ r_t+ \gamma Q_{\pi}(s_{t+1},a)|s_t,a_t\}.
\end{align}
 The optimal action-value function
$Q^{*}(s_t, a)$ is the cheat sheet sought by the RL agent, which
is defined as the maximum expectation of the cumulative
discounted return from state $s_t$:
\begin{equation}
 Q^{*}(s_t,a)= E_{s_{t+1}}\{ r_t+\gamma \max_{a} Q^*(s_t,a)|s_t,a\}.
\end{equation}

In DRL, a function approximation technique (a DNN in the considered case) is used to
learn a parametrized value function $Q(s, a;\theta)$ in order to approximate the optimal Q-values. The one-step
look ahead $r_t +  \gamma \max_a Q(s_{t+1}, a; \theta_q)$ is the target to obtain
$Q(s_t, a; \theta_q)$. Therefore the function $Q(s_t, a, \theta_q)$ is determined
by the parameters $\theta_q$. The selection of a good action relies on accurate action-value estimation, and thus DQN attempts to find the optimal parameters $\theta^*_q$  to minimize the
loss function:
\begin{equation}
 L({\theta_q})= (r_t +  \gamma \max_a Q(s_{t+1}, a; \theta_q)-Q(s_t, a; \theta_q))^2.
\end{equation}

Similar to classical Q-learning, the agent collects experiences by interacting with the environment. The network trainer constructs a data set $\mathcal{D}$ by collecting the experiences until time $t$ in the form of $(s_{t-1}, a_{t-1}, r_t, s_t)$. We optimize the loss function $L({\theta_q})$ using the collected data set $\mathcal{D}$. 

In the early stages of training, the agent estimation is not accurate, and a dynamic $\epsilon$-greedy policy is adopted to control the actions, where the agent with a certain probability explores different actions regardless of their reward. This strategy promotes accurate estimation over time, and reduces the risk of overfitting the model to actions with high rewards in the first phase of training. 

By substituting the DQN cost function into equation (\ref{eq:fed}), we obtain the FDQN cost as:
\begin{equation}
 \min_{\theta_q} L(\theta_q)=\sum_{i=1}^N w_i L_i(\theta_{q_i}),
\end{equation}
The overall learning procedure for FDQN is summarized in Algorithm 1.

\begin{algorithm}
    \caption{FDQN}
  \begin{algorithmic}[1]
    \item[\textbf{Input:}] Aggregation period $Ag$, learning rate $lr$, number of training episodes $N_e$, episode horizon time $T$, exploration parameter $\epsilon$, Initial $\theta_q$
    \STATE \textbf{Initialization} get initial $\theta_q$ from server
    \FOR{$e:=1$ to $N_e$}
    \STATE get initial state $S$
      \FOR{$t:=1$ to $T$}
        \STATE draw a random number $r \in [0,1]$
                \[ a_t =
                  \begin{cases}
                    \argmax_a Q(s_t,a,\theta_q^e)       & \quad \text{if } r> \epsilon\\
                    \text{pick uniformly action } a  & \quad \text{else }
                  \end{cases}
                \]
        \STATE Take action  $a_t$, go to state  $s_{t+1}$ and get reward $r_{t+1}$
        \STATE Store$\{a_t,s_t,r_{t+1},s_{t+1}\}$
        \ENDFOR
        \STATE update $\theta_q^{e+1}=\theta_q^{e}-lr \nabla_{\theta_q} L(\theta_q^e)$
      \IF{$e \mod{Ag=0}$} 
      \STATE send $\theta_q^t$ to server for aggregation
      \STATE get aggregated  $\theta_q^e$ from server
      \ENDIF
    \ENDFOR
    \item[\textbf{Output:}] $\theta_q^*$
  \end{algorithmic}
\end{algorithm}

\subsection{Federated Deep Policy Gradient}

In contrast to value based methods such as Q-learning, policy gradient algorithms directly optimize the policy without estimating the Q-values. This approach is more robust to the overestimation problem that affects value based methods. Using a DNN as a function approximator, we define the parametrized policy $\pi(a|s;\theta_{p})$, where  $\theta_{p}$ are the DNN weights. Herein, we focus specifically on Reinforce  as a policy gradient based learning algorithm that updates the policy based on the Monte-Carlo estimation of the agent average return. The objective of Reinforce is defined as:
\begin{equation}
J(\theta_p)=  E_{\pi} \{ \pi(a|s;\theta_{p}) R\},
\end{equation}
given the parameters $\theta_{p}$ and state $s$, the policy network generates a stochastic policy, that is, a probability vector over actions. Taking the gradient with respect to $\theta_p$ we obtain:
\begin{equation}
\nabla_{\theta_p} J(\theta_p)=  E_{\pi_{\theta_p}} \{\nabla_{\theta_p} \log ( \pi(a|s,\theta_p)) R\},
\end{equation}

\begin{table*}[tb]
\centering
{\small
\begin{tabular}{ l r r r r r r r r r }
\toprule
& \multicolumn{1}{c}{FDQN} & \multicolumn{1}{c}{DQN-Dist} & \multicolumn{1}{c}{DQN-Cent}  & \multicolumn{1}{c}{FDPG} & \multicolumn{1}{c}{DPG-Dist} & \multicolumn{1}{c}{DPG-Cent} & \multicolumn{1}{c}{WMMSE} & \multicolumn{1}{c}{Max Power}\\  
\midrule
\textbf{Mean} (bit/S/Hz) & 1.601  & 1.234 & 1.359 & 1.521 & 0.873 & 1.590 & 1.376 & 0.525  \\ 
\textbf{STD} (bit/S/Hz)  & 0.233 & 0.233 & 0.211 & 0.231  & 0.227  & 0.241 & 0.205 & 0.182 \\
\textbf{Average Execution Time} (S)  & 2.5 $\times 10^{-4}$ & 2.6 $\times 10^{-4}$ & 7.6 $\times 10^{-4}$ & 2.7$\times 10^{-4}$  & 2.7$\times 10^{-4}$  & 8.5 $\times 10^{-4}$ & 1.7$\times 10^{-2}$ & 9.2$\times 10^{-6}$ \\
\textbf{Communication Overhead }  & 0.01 & 0 & 1 & 0.01  & 0  & 1 & 1 & 0  \\
\bottomrule
\end{tabular}
}
\caption{Baselines and FDRL Performance Comparison}
\label{Table1:Comparisonbaseline}
\end{table*}
Parameters are updated to increase the probability of actions associated with higher rewards trajectories. 
Since Reinforce outputs a stochastic policy with non zeros probability over all actions, it does not require an exploration procedure such as the $\epsilon$-greedy strategy described earlier. At the test time, the optimal policy can be obtained by deterministically selecting the action with the largest probability.

By substituting the Reinforce cost function into equation (\ref{eq:fed}) we obtain the FDPG cost:
\begin{equation}
 \min_{\theta_p} J(\theta_p)=\sum_{i=1}^N w_i J_i(\theta_{p_i}).
\end{equation}
The overall training procedure for FDPG is summarized in Algorithm 2.

\begin{algorithm}
    \caption{FDPG}
  \begin{algorithmic}[1]
    \item[\textbf{Input:}]  Aggregation period $Ag$, learning rate $lr$, number of training episodes $N_e$, episode horizon time $T$, Initial $\theta_p$
    \STATE \textbf{Initialization} get initial $\theta_p$ from server
    \FOR{$e:=1$ to $Ne$}
    \STATE get initial state $S$
      \FOR{$t:=1$ to $T$}
        \STATE sample action $a_t$ from distribution $\pi(a_t|s_t,\theta_p^e)$
        \STATE Take action  $a_t$, go to state  $s_{t+1}$ and get reward $r_{t+1}$
        \ENDFOR
        \STATE update $\theta_p^{e+1}=\theta_p^{e}+lr \nabla_{\theta_p} J(\theta_p^e)$
      \IF{ $e \mod{Ag=0}$} 
      \STATE send $\theta_p^e$ to server for aggregation
      \STATE get aggregated  $\theta_p^e$ from server
      \ENDIF
    \ENDFOR
    \item[\textbf{Output:}] $\theta_p^*$
  \end{algorithmic}
\end{algorithm}

\vspace{12pt}

\section{Results}

In this section, we provide a thorough performance evaluation of the proposed federated algorithms. We implement all the models in PyTorch and use the Adam optimizer to train them, setting the learning rate to $lr=0.001$. We use the same network architecture for DQL and deep policy gradient (Reinforce). We employ a neural network whose number of inputs are equal to the state dimension. The network has two hidden layers with 128 and 64 neurons respectively, followed by Relu activation functions. The output dimension is set to be the same as the number of  discretized power levels ($M=10$ in the results) for policy gradient and DQN. The models will be released in our GitHub repository\footnote{https://github.com/PeymanTehrani/FDRL-PC-Dyspan}.
In the training procedure, we consider a multi cell network with $N = 25$ cells in which in each cell serves $K=4$ users. The Doppler frequency
and time slot period are set to $f_d = 10$Hz and $T_s = 20$ms respectively.
We also consider the maximum transmission power to be $P_{max} = 38$dbm and the control parameter in reward function $\beta = 1$.

\begin{figure}
\centering
        \includegraphics[width=.8\columnwidth,height=.5\columnwidth]{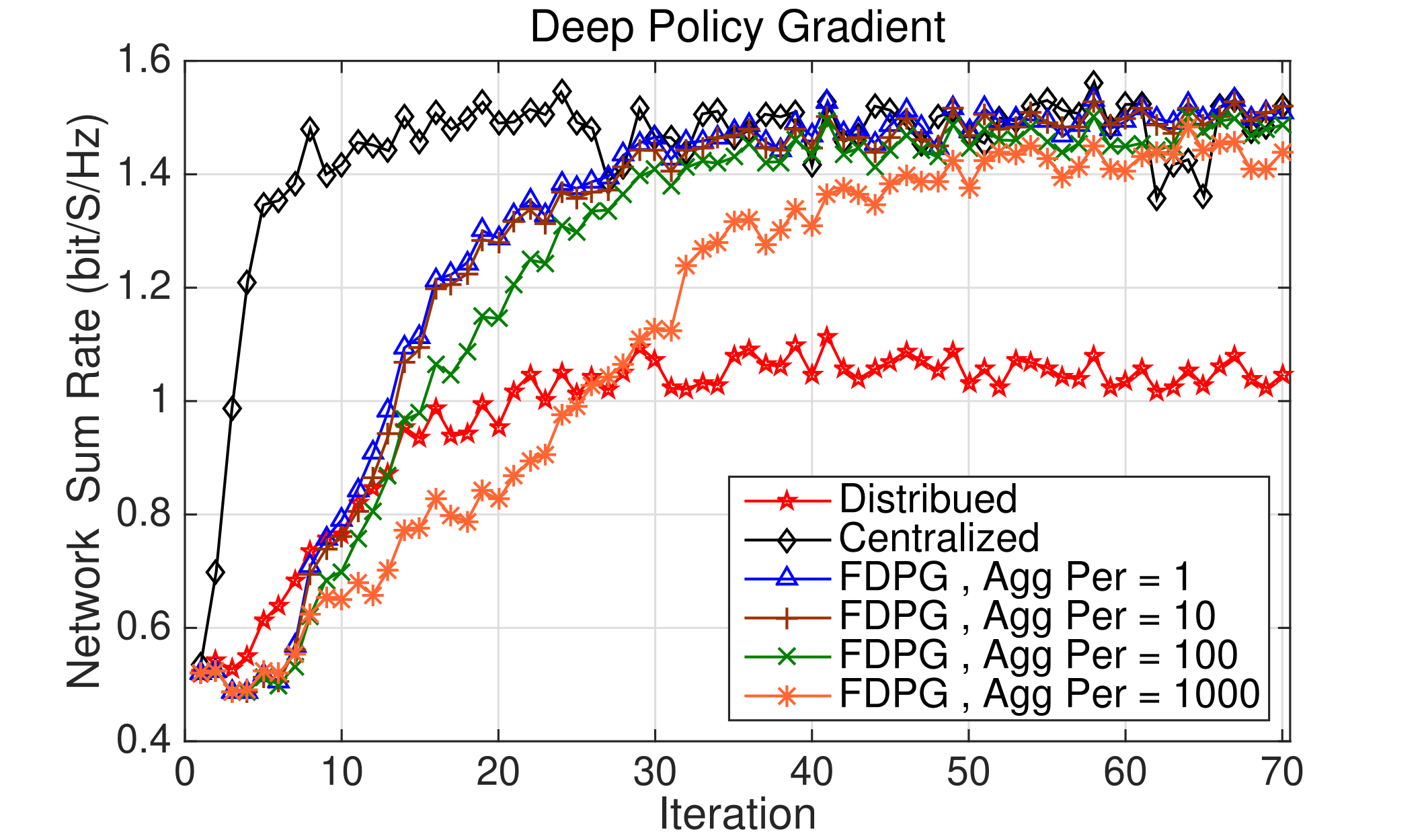}
        \caption{Comparison of the convergence of distributed deep policy gradients with its federated implementation for different aggregation frequencies.}
        \label{fig:FLpolicy}
\end{figure}

\begin{figure}
\centering
        \includegraphics[width=.8\columnwidth,height=.5\columnwidth]{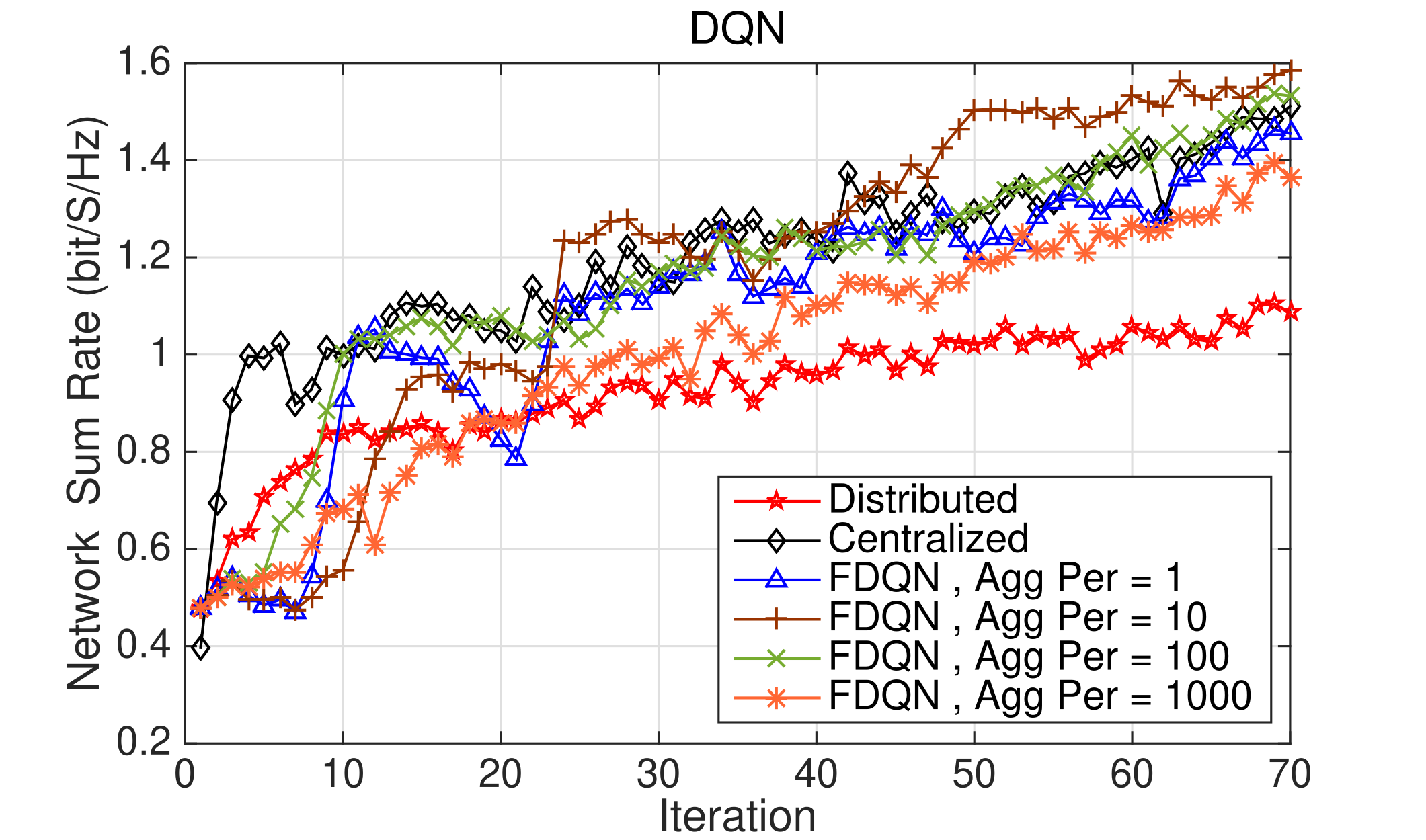}
        \caption{Comparison of the convergence of distributed deep Q-Learning  with its federated implementation for different aggregation frequencies.}
        \label{fig:FL_DQN}
\end{figure}

Figs \ref{fig:FLpolicy} and \ref{fig:FL_DQN}  depict the average per user rate in the network for centralized, distributed and federated implementations of DPG and DQN algorithms with different aggregation periods. We train each model over 7000 episodes, where each episode itself contains a horizon of $T{=}10$ time slots. To plot smoother curves, the results of groups of $100$ iterations is reduced to its average in the plots. In the distributed implementation, it is assumed each BS only trains its own model based on the its observed states and does not share the model weights with any other BSs (Agents) or the central server while in the centralized case in each episode all BSs send their user states to the central server in order to learn a global model.


Comparing the results of distributed and federated algorithms, we can see that the achieved rate per user improves significantly in the latter case. Additionally, when comparing the centralized case with federated version we can observe that they have similar performance, but the federated approach the communication between the BSs and server is up to $0.001$ times smaller (in $Agg Per = 1000$ case).  Furthermore, we note that a higher Aggregation frequency leads to faster convergence speed for a penalty of having larger communication overhead. It can be observed that FDPG convergence is smoother compared to FDQN for all the different aggregation frequencies, thus providing more coherent performance to users. This can be explained by the use of the $\epsilon$-greedy approach for exploration, which increases the random behavior until convergence is reached, while in the policy gradient-based algorithms, the gradient update selects state-action trajectories associated with higher average rewards which could benefits from model aggregation.

In Table \ref{Table1:Comparisonbaseline} we compare the different implementations of  DRL (federated, distributed and centralized) alongside with other baselines such as weighted minimum mean squared error (WMMSE) \cite{shi2011iteratively} and maximum  power transmission (Max Power). Here, we tested all the models on 1000 different random episodes and reported the mean and standard deviation of the network sum rate, average execution time  and the server-BS communication overhead for each of these algorithms. For the F-DRL we set the aggregation period to be equal to $100$. Based on the results, it is apparent that the proposed F-DRL approach provides a good balance between high throughput, fast execution time and low overhead. This proves that incorporating the F-DRL strategy at the nextG wireless networks leads to fast response, high performance and efficient automated networks.

\begin{figure}
\centering
        \includegraphics[width=.8\columnwidth,height=.42\columnwidth]{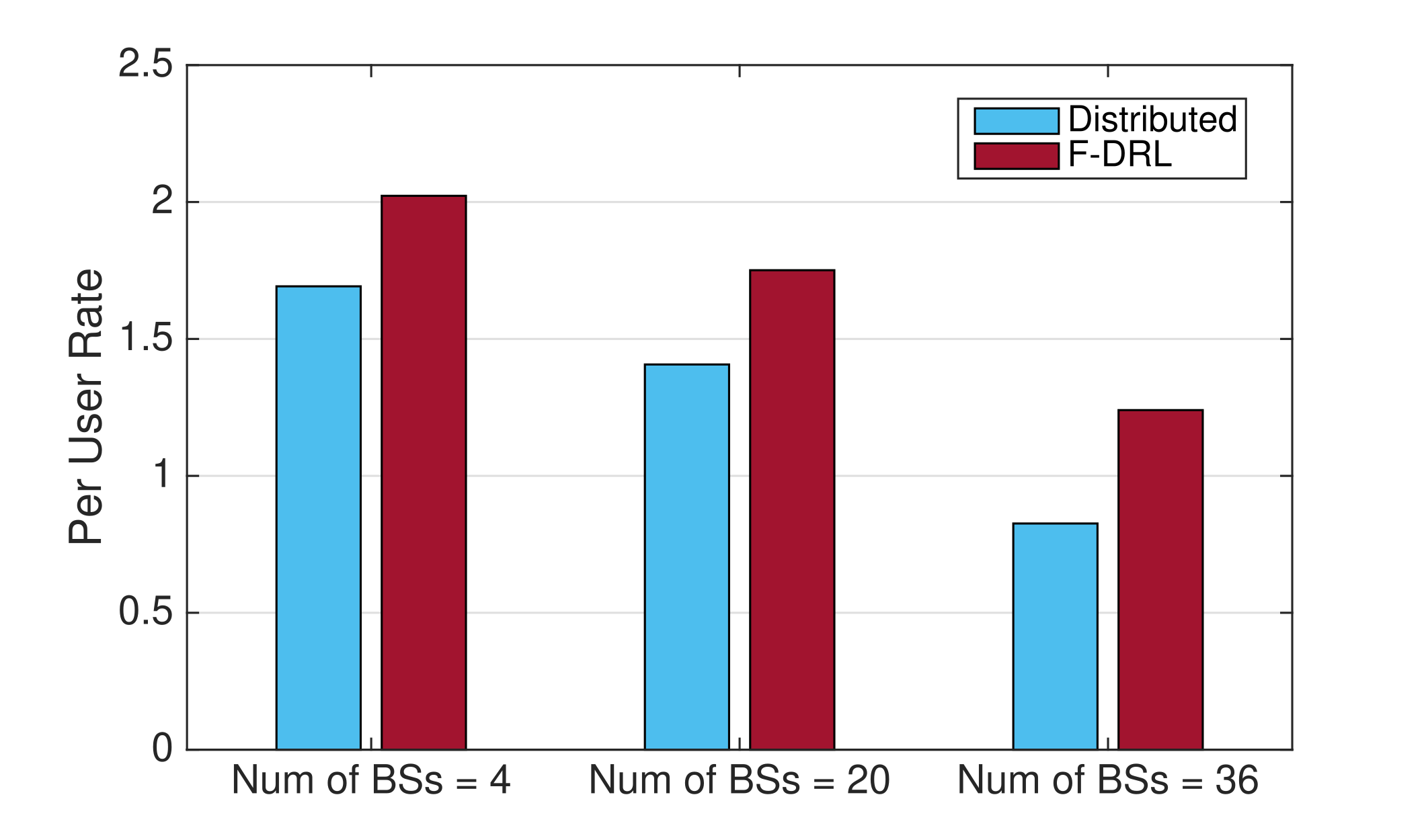}
        \caption{Performance Comparison between the federated and distributed deep policy gradient for different number of available cells.}
        \label{fig:BS_policy_bar}
\end{figure}

In Fig. \ref{fig:BS_policy_bar}, the performance of the distributed and federated deep policy gradient algorithms are compared in terms of average per user rate. We can see that as the size of the network increases, the overall interference increases and consequently the average data rate per user decreases. Notably, when the number of BSs is very large (\emph{e.g.}, $36$) the gain of federated frameworks over fully distributed optimization is about $40\%$, while for smaller networks the gain is around $15\%$.

\section{Conclusions}
In this paper, we proposed federated deep reinforcement learning as a tool to solve a distributed power control problem in a wireless multi-cell network. We investigated the performance of DRL with value based and policy based methods and compared their federated version and distributed version implementations. We demonstrated by simulation that aggregating the models of the BSs can improve the performance in terms of overall network sum rate compared to fully distributed implementation, while also being more bandwidth efficient comparing to fully centralized scenarios. We also showed that the F-DRL approach outperforms the conventional optimization-based baselines both in terms of performance and execution time.

\bibliographystyle{IEEEtran}
\bibliography{ref}

\begin{thebibliography}{10}
\providecommand{\url}[1]{#1}
\csname url@samestyle\endcsname
\providecommand{\newblock}{\relax}
\providecommand{\bibinfo}[2]{#2}
\providecommand{\BIBentrySTDinterwordspacing}{\spaceskip=0pt\relax}
\providecommand{\BIBentryALTinterwordstretchfactor}{4}
\providecommand{\BIBentryALTinterwordspacing}{\spaceskip=\fontdimen2\font plus
\BIBentryALTinterwordstretchfactor\fontdimen3\font minus
  \fontdimen4\font\relax}
\providecommand{\BIBforeignlanguage}[2]{{%
\expandafter\ifx\csname l@#1\endcsname\relax
\typeout{** WARNING: IEEEtran.bst: No hyphenation pattern has been}%
\typeout{** loaded for the language `#1'. Using the pattern for}%
\typeout{** the default language instead.}%
\else
\language=\csname l@#1\endcsname
\fi
#2}}
\providecommand{\BIBdecl}{\relax}
\BIBdecl

\bibitem{tehrani2016resource}
P.~Tehrani, F.~Lahouti, and M.~Zorzi, ``Resource allocation in ofdma networks
  with half-duplex and imperfect full-duplex users,'' in \emph{2016 IEEE
  international conference on communications (ICC)}.\hskip 1em plus 0.5em minus
  0.4em\relax IEEE, 2016.

\bibitem{mnih2015human}
V.~Mnih, K.~Kavukcuoglu, D.~Silver, A.~A. Rusu, J.~Veness, M.~G. Bellemare,
  A.~Graves, M.~Riedmiller, A.~K. Fidjeland, G.~Ostrovski \emph{et~al.},
  ``Human-level control through deep reinforcement learning,'' \emph{nature},
  vol. 518, no. 7540, pp. 529--533, 2015.

\bibitem{khan2020centralized}
A.~A. Khan and R.~Adve, ``Centralized \& distributed deep reinforcement
  learning methods for downlink sum-rate optimization,'' \emph{IEEE
  Transactions on Wireless Communications}, 2020.

\bibitem{meng2020power}
F.~Meng, P.~Chen, L.~Wu, and J.~Cheng, ``Power allocation in multi-user
  cellular networks: Deep reinforcement learning approaches,'' \emph{IEEE
  Transactions on Wireless Communications}, 2020.

\bibitem{sinan2020deep}
Y.~Sinan~Nasir and D.~Guo, ``Deep actor-critic learning for distributed power
  control in wireless mobile networks,'' \emph{arXiv e-prints}, pp.
  arXiv--2009, 2020.

\bibitem{zhang2019calibrated}
X.~Zhang, M.~R. Nakhai, G.~Zheng, S.~Lambotharan, and B.~Ottersten,
  ``Calibrated learning for online distributed power allocation in small-cell
  networks,'' \emph{IEEE Transactions on Communications}, 2019.

\bibitem{mcmahan2017communication}
B.~McMahan, E.~Moore, D.~Ramage, S.~Hampson, and B.~A. y~Arcas,
  ``Communication-efficient learning of deep networks from decentralized
  data,'' in \emph{Artificial Intelligence and Statistics}.\hskip 1em plus
  0.5em minus 0.4em\relax PMLR, 2017.

\bibitem{kumar2017federated}
S.~Kumar, P.~Shah, D.~Hakkani-Tur, and L.~Heck, ``Federated control with
  hierarchical multi-agent deep reinforcement learning,'' \emph{arXiv preprint
  arXiv:1712.08266}, 2017.

\bibitem{wang2020attention}
X.~Wang, R.~Li, C.~Wang, X.~Li, T.~Taleb, and V.~C. Leung, ``Attention-weighted
  federated deep reinforcement learning for device-to-device assisted
  heterogeneous collaborative edge caching,'' \emph{IEEE Journal on Selected
  Areas in Communications}, vol.~39, no.~1, pp. 154--169, 2020.

\bibitem{wang2020federated}
X.~Wang, C.~Wang, X.~Li, V.~C. Leung, and T.~Taleb, ``Federated deep
  reinforcement learning for internet of things with decentralized cooperative
  edge caching,'' \emph{IEEE Internet of Things Journal}, 2020.

\bibitem{zhu2020federated}
Z.~Zhu, S.~Wan, P.~Fan, and K.~B. Letaief, ``Federated multi-agent actor-critic
  learning for age sensitive mobile edge computing,'' \emph{arXiv preprint
  arXiv:2012.14137}, 2020.

\bibitem{ren2019federated}
J.~Ren, H.~Wang, T.~Hou, S.~Zheng, and C.~Tang, ``Federated learning-based
  computation offloading optimization in edge computing-supported internet of
  things,'' \emph{IEEE Access}, vol.~7, pp. 69\,194--69\,201, 2019.

\bibitem{nguyen2021federated}
D.~C. Nguyen, M.~Ding, P.~N. Pathirana, A.~Seneviratne, J.~Li, D.~Niyato, and
  H.~V. Poor, ``Federated learning for industrial internet of things in future
  industries,'' \emph{arXiv preprint arXiv:2105.14659}, 2021.

\bibitem{zhang2018power}
Y.~Zhang, C.~Kang, T.~Ma, Y.~Teng, and D.~Guo, ``Power allocation in multi-cell
  networks using deep reinforcement learning,'' in \emph{2018 IEEE 88th
  Vehicular Technology Conference (VTC-Fall)}.\hskip 1em plus 0.5em minus
  0.4em\relax IEEE, 2018.

\bibitem{meng2019power}
F.~Meng, P.~Chen, and L.~Wu, ``Power allocation in multi-user cellular networks
  with deep q learning approach,'' in \emph{IEEE International Conference on
  Communications (ICC)}.\hskip 1em plus 0.5em minus 0.4em\relax IEEE, 2019, pp.
  1--6.

\bibitem{nasir2019multi}
Y.~S. Nasir and D.~Guo, ``Multi-agent deep reinforcement learning for dynamic
  power allocation in wireless networks,'' \emph{IEEE Journal on Selected Areas
  in Communications}, vol.~37, no.~10, 2019.

\bibitem{ding2020deep}
H.~Ding, F.~Zhao, J.~Tian, D.~Li, and H.~Zhang, ``A deep reinforcement learning
  for user association and power control in heterogeneous networks,'' \emph{Ad
  Hoc Networks}, vol. 102, p. 102069, 2020.

\bibitem{saeidian2020downlink}
S.~Saeidian, S.~Tayamon, and E.~Ghadimi, ``Downlink power control in dense 5g
  radio access networks through deep reinforcement learning,'' in \emph{IEEE
  International Conference on Communications (ICC)}, 2020.

\bibitem{lei2020deep}
W.~Lei, Y.~Ye, and M.~Xiao, ``Deep reinforcement learning based spectrum
  allocation in integrated access and backhaul networks,'' \emph{IEEE
  Transactions on Cognitive Communications and Networking}, 2020.

\bibitem{tan2020deep}
J.~Tan, Y.-C. Liang, L.~Zhang, and G.~Feng, ``Deep reinforcement learning for
  joint channel selection and power control in d2d networks,'' \emph{IEEE
  Transactions on Wireless Communications}, 2020.

\bibitem{shi2011iteratively}
Q.~Shi, M.~Razaviyayn, Z.-Q. Luo, and C.~He, ``An iteratively weighted mmse
  approach to distributed sum-utility maximization for a mimo interfering
  broadcast channel,'' \emph{IEEE Transactions on Signal Processing}, vol.~59,
  no.~9, pp. 4331--4340, 2011.

\end{thebibliography}
\end{document}